\documentclass[runningheads]{llncs}
\usepackage{graphicx}
\usepackage{amsmath,amssymb} 
\usepackage{color}
\usepackage{multirow} 
\usepackage[ruled]{algorithm2e}
\usepackage{algorithmic}
\usepackage{multicol}
\usepackage{hhline}
\usepackage{pifont}

\begin{document}
\title{Facial Dynamics Interpreter Network: What are the Important Relations between Local Dynamics for Facial Trait Estimation?} 

\titlerunning{Facial Dynamics Interpreter Network}
%
\author{Seong Tae Kim \and Yong Man Ro{*}}
%
\authorrunning{S.T. Kim and Y.M. Ro}
%

\institute{School of Electrical Engineering, KAIST, Daejeon, Republic of Korea \\
	\email{\{stkim4978,ymro\}@kaist.ac.kr}\\}
\maketitle              
\begin{abstract}
	Human face analysis is an important task in computer vision. According to cognitive-psychological studies, facial dynamics could provide crucial cues for face analysis. The motion of a facial local region in facial expression is related to the motion of other facial local regions. In this paper, a novel deep learning approach, named facial dynamics interpreter network, has been proposed to interpret the important relations between local dynamics for estimating facial traits from expression sequence. The facial dynamics interpreter network is designed to be able to encode a relational importance, which is used for interpreting the relation between facial local dynamics and estimating facial traits. By comparative experiments, the effectiveness of the proposed method has been verified. The important relations between facial local dynamics are investigated by the proposed facial dynamics interpreter network in gender classification and age estimation. Moreover, experimental results show that the proposed method outperforms the state-of-the-art methods in gender classification and age estimation. 
\end{abstract}

\section{Introduction}
Analysis of human face has been an important task in computer vision because it plays a major role in soft biometrics, and human-computer interaction \cite{reid2013soft,dantcheva2016else}. Facial behavior is known to benefit perception of the identity \cite{roark2003psychological,pilz2006search}. In particular, facial dynamics play crucial roles for improving the accuracy of facial trait estimation such as age estimation or gender classification \cite{dibekliouglu2015combining,dantcheva2017gender}.

In recent progress of deep learning, convolutional neural networks (CNN) have shown outstanding performance on many fields of computer vision. Several research efforts have been devoted to developing spatio-temporal feature representation in various applications such as action recognition \cite{ji20133d,tran2015learning,donahue2015long,karpathy2014large} and activity parsing \cite{yue2015beyond,lea2016segmental}. In \cite{karpathy2014large}, a long short-term memory (LSTM) network has been designed on top of CNN features to encode dynamics in video. The LSTM network is a variant of recurrent neural network (RNN), which is designed to capture long-term temporal information in sequential data \cite{hochreiter1997long}. By using the LSTM, the temporal correlation of CNN features was effectively encoded.

Recently, a few research efforts have been made regarding facial dynamic feature encoding for a facial analysis \cite{dibekliouglu2015combining,kim2016facial,dantcheva2017gender,kim2017multi}.It is generally known that the dynamic features of local regions are valuable for facial trait estimation\cite{dibekliouglu2015combining,dantcheva2017gender}. Usually, the motion of facial local region in facial expression is related to the motion of other facial regions \cite{tong2007facial,zhao2016joint}. However, to the best of our knowledge, there are no studies that utilize relations between facial motions and interpret the important relations between local dynamics for facial trait estimation.

In this paper, a novel deep network has been proposed for interpreting relations between local dynamics in facial trait estimation. To interpret the relations between facial local dynamics, the proposed deep network consists of a facial local dynamic feature encoding network and a facial dynamics interpreter network. By the facial dynamics interpreter network, the importance of relations for estimating facial traits is encoded. The main contributions of this study are summarized in following three aspects:
\begin{enumerate}
	\item We propose a novel deep network which estimates facial traits by using relations between facial local dynamics of smile expression . 
	\item The proposed deep network has been designed to be able to interpret the relations between local dynamics in facial trait estimation. For that purpose, the relational importance is devised. The relational importance is encoded from the relational features of facial local dynamics. The relational importance is used for interpretation of important relations in facial trait estimation. 
	\item To validate the effectiveness of the proposed method, comparative experiments have been conducted on two facial trait estimation problems (\textit{i.e.} age estimation and gender classification). In the proposed method, the facial trait estimation is conducted by combining the relational features based on the relational importance. By exploiting the relational features and considering the importance of relations, the proposed method could more accurately estimate facial traits compared with the state-of-the-art methods. 
\end{enumerate}

\section{Related Work}

\noindent
\textbf{Age Estimation and Gender Classification.}  A lot of research efforts have been devoted to development of automatic age estimation and gender classification techniques from face image \cite{guo2009human,alnajar2012learning,toews2009detection,makinen2008evaluation,bekios2014robust,juefei2016deepgender,li2015shape,uvrivcavr2016structured,toews2009detection,makinen2008evaluation,bekios2014robust}. 
Recently, deep learning methods show notable potential in various face analysis tasks. One of the main focus of these methods is to design suitable deep network structure for some specific tasks. Parkhi et al. \cite{parkhi2015deep} reported VGG-style CNN learned from large-scale static face images. Deep learning based age estimation method and gender classification method have been reported but they were mostly designed on static face image \cite{juefei2016deepgender,li2015shape,uvrivcavr2016structured,levi2015age}.

\noindent
\textbf{Facial Dynamic Analysis.}  The temporal dynamics of face have been ignored in both age estimation and gender classification. Recent studies have reported that facial dynamics could be an important cue for facial trait estimation \cite{hadid2011analyzing,demirkus2010gender,dibekliouglu2012smile,dibekliouglu2015combining,dantcheva2017gender}. With aging, the face loses muscle tone and underlying fat tissue, which creates wrinkles, sunken eyes and increases crow’s feet around the eyes \cite{dibekliouglu2015combining}. Aging also affects facial dynamics along with appearance. As a human being gets older, the elastic fibers of the face show fraying. Therefore facial dynamic features of local facial regions are important cues for age estimation. In cognitive-psychological studies \cite{cashdan1998smiles,hess2004facial,simon2004gender,adams2015intersection}, evidence for gender-dimorphism in the human expression has been reported. Females express emotions more frequently compared with males. Males have a tendency to show restricted emotions and to be unwilling to self-disclose intimate feelings \cite{dantcheva2017gender}. In \cite{dantcheva2017gender}, Dantcheva et al. used dynamic descriptors extracted from facial landmarks for gender classification. However, there are no studies for learning relations of dynamic features in facial trait estimation.

\noindent
\textbf{Relational Network.}  In this paper, we propose a novel deep learning architecture for analyzing relations of facial dynamic features in facial trait estimation. A relational network has been reported in in visual question and answering (VQA) \cite{santoro2017simple}. In \cite{santoro2017simple}, the authors defined an object as a neuron on feature map obtained from CNN and designed a neural network for relational reasoning. However, it was designed for image-based VQA. In this paper, the proposed method automatically encodes the importance of relations by considering the locational information on face. By utilizing the importance of relations, the proposed method could interpret the relations between facial dynamics in facial trait estimation.

\section{Proposed Facial Dynamics Interpreter Network}

\begin{figure}[!t]
	\centering
	\includegraphics[width=1\linewidth] {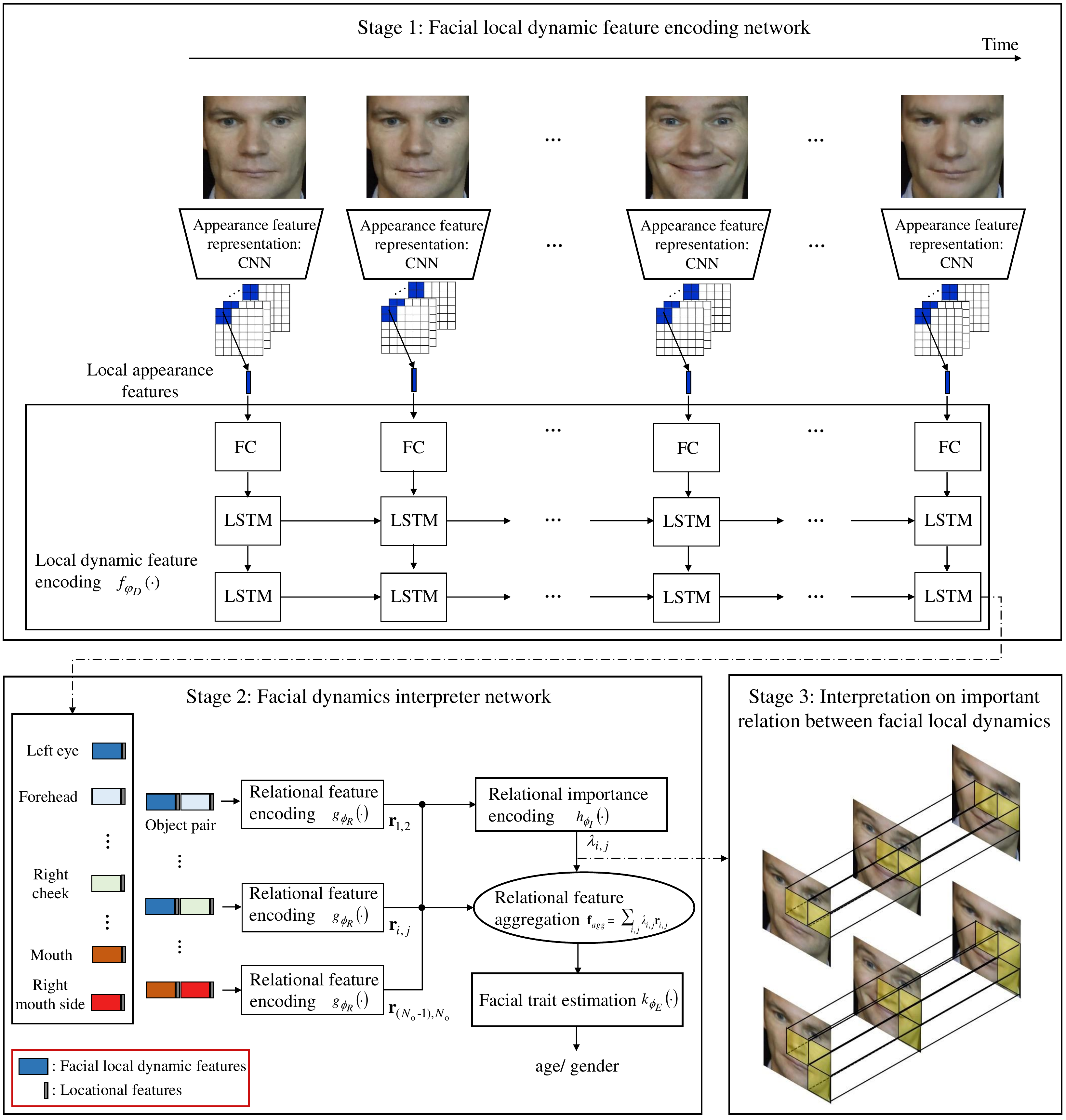}
	\caption{Overall structure of the proposed facial dynamics interpreter network.}
	\label{fig1}
\end{figure}

Overall structure of the proposed facial dynamics interpreter network is shown in Fig.\ref{fig1}. The aim of the proposed method is to interpret the important relations between local dynamics in facial trait estimation from expression sequence. The proposed method largely consists of the facial local dynamic feature encoding network, the facial dynamics interpreter network, and interpretation on important relation between facial local dynamics. The details are described in the following subsections.

\subsection{Facial Local Dynamic Feature Encoding Network}
Given a face sequence, appearance features are computed by CNN on each frame. For the purpose of appearance feature extraction, we employ the VGG-face network \cite{parkhi2015deep} which is trained with large-scale face images. The pre-trained VGG face model is used to get off-the-shelf CNN features in this study. With given CNN features, the proposed facial dynamics interpreter network has been investigated. The output of convolutional layer in the VGG-face network is used as feature map of facial appearance representation.

Based on the feature map, the face is divided into $N_0$ local regions. The location of local regions was determined to interpret the relation of local dynamics base on semantically meaningful facial local region (\textit{i.e.} left eye, forehead, right eye, left cheek, nose, right cheek, left mouth side, mouth, and right mouth side in this study). Note that each face sequence is automatically aligned based on the landmark detection \cite{asthana2014incremental}. Let $\mathbf{x}_i^t$ denote the local appearance features of \textit{i}-th facial local part at \textit{t}-th time step. To encode local dynamic features, an LSTM network has been devised with fully-connected layer on top of the local appearance features $\mathbf{X}_i=\left\{\mathbf{x}_i^1,...,\mathbf{x}_i^t,...,\mathbf{x}_i^T\right\}$ as followings:
\begin{equation}\label{eq1}
\begin{aligned}
\mathbf{d}_i=f_{\phi_{D}}(\mathbf{X}_i),
\end{aligned} 
\end{equation}
where  $\mathbf{d}_i$ denotes the facial local dynamic feature of \textit{i}-th local part and $f_{\phi_{D}}$ is a function with learnable parameters $\phi_{D}$. $f_{\phi_{D}}$ consists of the fully-connected layer and the LSTM layers as shown in Fig. 1. \textit{T} denotes the length of face sequence. The LSTM network could deal with the different length of sequences. The various dynamic related features including variation of appearance, amplitude, speed, and acceleration could be encoded from the sequence of local appearance features. The detailed configuration of the network used in the experiments will be presented in Section 4.1.

\subsection{Facial Dynamics Interpreter Network}
We extract object features (\textit{i.e.} facial local dynamic features and locational features) for pairs of objects. The locational features are defined as the central position of the object (\textit{i.e.} facial local region). For the purpose of telling the location information of objects to the facial dynamics interpreter network, the local dynamic features and the locational features are embedded and defined as object features $\mathbf{o}_i$. The object feature can be written as
\begin{equation}\label{eq2}
\begin{aligned}
\mathbf{o}_i=[\mathbf{d}_i,p_i,q_i],
\end{aligned} 
\end{equation}
where $[p_i,q_i]$ denotes the normalized central position of \textit{i}-th object. 

The design philosophy of the proposed facial dynamics interpreter network is to make the functional form of a neural network which captures the core relations for facial trait estimation. The importance of the relation could be different for each pair of object features. The proposed facial dynamics interpreter network is designed to encode relational importance in facial trait estimation. The relational importance could be used for interpreting the relation between local dynamics in facial trait estimation.

Let $\lambda_{i,j}$ denote a relational importance between \textit{i}-th and \textit{j}-th object feature. The relational feature, which represents latent relation between two objects for facial trait estimation, can be written as
\begin{equation}\label{eq3}
\begin{aligned}
\mathbf{r}_{i,j}=g_{\phi_{R}}(\mathbf{s}_{i,j}),
\end{aligned} 
\end{equation}
where $g_{\phi_{R}}$ is a function with learnable parameters $\phi_{R}$. $\mathbf{s}_{i,j}=(\mathbf{o}_i,\mathbf{o}_j)$ is relation pair from \textit{i}-th and \textit{j}-th facial local parts. $\mathbf{S}=\left\{\mathbf{s}_{1,2},\cdots,\mathbf{s}_{i,j},\cdots,\mathbf{s}_{(N_0-1),N_0}\right\}$ is a set of relation pairs where $N_0$ denotes the number of objects in face. $\mathbf{o}_i$ and $\mathbf{o}_j$ denote the \textit{i}-th and \textit{j}-th object features, respectively. The relational importance $\lambda_{i,j}$ for relation between two object features ($\mathbf{o}_i$, $\mathbf{o}_j$) is encoded as: 
\begin{equation}\label{eq4}
\begin{aligned}
\lambda_{i,j}=h_{\phi_{I}}(\mathbf{r}_{i,j}),
\end{aligned} 
\end{equation}
where $h_{\phi_{I}}$ is a function with learnable parameters $\phi_{I}$. In this paper, $h_{\phi_{I}}$ is defined with $\phi_{I}=\left\{\left(\mathbf{W}_{1,2},\mathbf{b}_{1,2}\right),\cdots,\left(\mathbf{W}_{(N_0-1),N_0},\mathbf{b}_{(N_0-1),N_0}\right)\right\} $ as followings: 
\begin{equation}\label{eq5}
\begin{aligned}
h_{\phi_{I}}\left(\mathbf{r}_{i,j}\right)=\frac{\exp\left(\mathbf{W}_{i,j}\mathbf{r}_{i,j}+\mathbf{b}_{i,j}\right)}{\sum_{i,j}\exp\left(\mathbf{W}_{i,j}\mathbf{r}_{i,j}+\mathbf{b}_{i,j}\right)}.
\end{aligned} 
\end{equation}
The aggregated relational features $\mathbf{f}_{agg}$ are represented by
\begin{equation}\label{eq6}
\begin{aligned}
\mathbf{f}_{agg}=\sum_{i,j}\lambda_{i,j}\mathbf{r}_{i,j}.
\end{aligned} 
\end{equation}
Finally, the facial trait estimation can be performed with 
\begin{equation}\label{eq7}
\begin{aligned}
\mathbf{y}=k_{\phi_{E}}(\mathbf{f}_{agg}),
\end{aligned} 
\end{equation}
where $\mathbf{y}$ denotes estimated result and $k_{\phi_{E}}$ is a function with parameters $\phi_{E}$. $k_{\phi_{E}}$ and $g_{\phi_{R}}$ are implemented by by fully-connected layers.

\subsection{Interpretation on Important Relations between Facial Local Dynamics}
\begin{algorithm}[!t]
	\caption{Calculating relational importance of $N_I$ objects}
	\label{Alg:1}
	\textbf{Input} \\
	$\boldsymbol {\lambda}$: the set of relational importance of two facial local parts $\left\{\lambda_{1,2},\cdots,\lambda_{i,j},\cdots,\lambda_{(N_0-1),N_0}\right\}$\\
	$N_I$: the number of objects for interpretation  \\
	\textbf{Output:}  $\boldsymbol {\chi}$: Relational importance of $N_I$ objects   \\
	
	Let $N_R$ denote total number of relations with $N_I$ objects 
	$N_R=_{N_o}$C$_{N_I}$ \\
	
	\For{P from 1 to $N_R$ } {
		Let $\chi_P^{N_I}$ denote \textit{P}-th relational importance of ${N_I}$ object features ( \textbf{o}$_{p_1}$,\textbf{o}$_{p_2},\cdots,$\textbf{o}$_{p_{(N_I-1)}}$,\textbf{o}$_{p_{N_I}}$) \\
		Compute $\chi_P^{N_I}$ \\
		\quad \quad $\chi_P^{N_I}=\lambda_{p_1,p_2}+\cdots+\lambda_{p_1,p_{N_I}}+\cdots+\lambda_{p_{(N_I-1)},p_{N_I}}$ \\
	}
	$\boldsymbol {\chi}=\left\{\chi_1^{N_I},\cdots,\chi_P^{N_I},\cdots,\chi_{N_R}^{N_I}\right\}$
\end{algorithm}

The proposed method is useful for interpreting the relations in facial trait estimation. The relational importance calculated in Eq. (4) is utilized to interpret the relations of facial local dynamics. Note that the high relational importance values mean that the relational features of corresponding facial local parts are important for estimating facial traits. The pseudocodes for calculating relational importance of $N_I$ objects are given in Algorithm 1. By analyzing the relational importance, important relations for estimating facial traits could be explained. In Section 4.2 and 4.3, we discuss the important relations for age estimation and gender classification, respectively.

\section{Experiments}
\label{sec:Experiments}
\subsection{Experimental Settings}

\noindent
\textbf{Database.}
To evaluate the effectiveness of the proposed facial dynamics interpreter network, comparative experiments were conducted. For generalization purpose, we verified the proposed method on both age estimation and gender classification tasks. Age and gender were known as representative facial traits \cite{li2015shape}.  The public UvA-NEMO Smile database was used for both tasks \cite{dibekliouglu2012smile,dibekliouglu2012you}. The UvA-NEMO smile database has been known as the largest smile database \cite{dibekliouglu2015recognition}. The database consists of 1,240 smile videos collected from 400 subjects. Among 400 subjects, 185 subjects are female and remaining 215 subjects are male. The ages of subjects range from 8 to 76 years. For evaluating the performance of age estimation, we used the experimental protocol defined in \cite{dibekliouglu2015combining,dibekliouglu2012smile,dibekliouglu2012you}. The 10-fold cross-validation scheme was used to calculate the performance of the proposed method. Each fold was divided in a way where there was no subject overlap \cite{dibekliouglu2015combining,dibekliouglu2012smile,dibekliouglu2012you}. Each time an independent test fold was separated and it was only used for calculating the perofrmance. The remaining 9-folds were used to train the deep network and optimize hyper-parameters. To evaluate the performance of gender classification, we followed the experimental protocol used in \cite{dantcheva2017gender}.

\noindent
\textbf{Evaluation Metric.}
For age estimation, the mean absolute error (MAE) \cite{uvrivcavr2016structured} was utilized for evaluation. The MAE could measure the error between the predicted age and the ground-truth. The MAE was computed as follows:
\begin{equation}\label{eq8}
\begin{aligned}
\epsilon=\frac{{\sum_{n=1}^{N_{test}}\lVert\mathbf{\hat{y}}_n-\mathbf{y}_n^*\rVert}_1}{N_{test}},
\end{aligned} 
\end{equation}
where $\mathbf{\hat{y}}_n$ and $\mathbf{y}_n^*$ denote predicted age and ground-truth age of \textit{n}-th test sample, respectively. $N_{test}$ denotes the number of the test samples. For the case of gender classification, classification accuracy was used for evaluation. We reported the MAE and classification accuracy averaged over all test folds.

\noindent
\textbf{Implementation Details.}
The face images used in the experiments were automatically aligned based on the two eye locations detected by the facial landmark detection  \cite{asthana2014incremental}. The face images were cropped and resized to 96$\times$96 pixels. For the appearance representation, the frontal 10 convolutional layers and 4 max-pooling layers of VGG-face network was used. As a result, 6$\times$6$\times$512 size of feature map was obtained from each face image. Each facial local region was defined on the feature map with size of 2$\times$2$\times$512. In other words, there were 9 objects in face sequence ($N_0=9$). The fully-connected layer with 1024 units and the stacked LSTM layers were used for $f_{\phi_{D}}$. We stacked two LSTMs and each LSTM had 1024 memory cells. Two-layer full-connected layers consisting of 4096 units (with dropout \cite{srivastava2014dropout}) per layer was used for $g_{\phi_{R}}$ with RELU \cite{nair2010rectified}. $h_{\phi_{I}}$ was implemented by a fully-connected layer and softmax function. Two-layer fully-connected layers consisting of 2048, 1024 units (with dropout, RELU, and batch normalization \cite{ioffe2015batch}) and one fully-connected layer (1 neuron for age estimation and 2 neurons for gender classification) were used for $k_{\phi_{E}}$. The mean squared error was used for training the deep network in age estimation. The cross-entropy loss was used for training the deep network in gender classification.

\subsection{Age Estimation}

\subsubsection{Interpreting Relations between Facial Local Dynamics in Age Estimation.}
\begin{figure}[!t]
	\centering
	\includegraphics[width=4.8in]{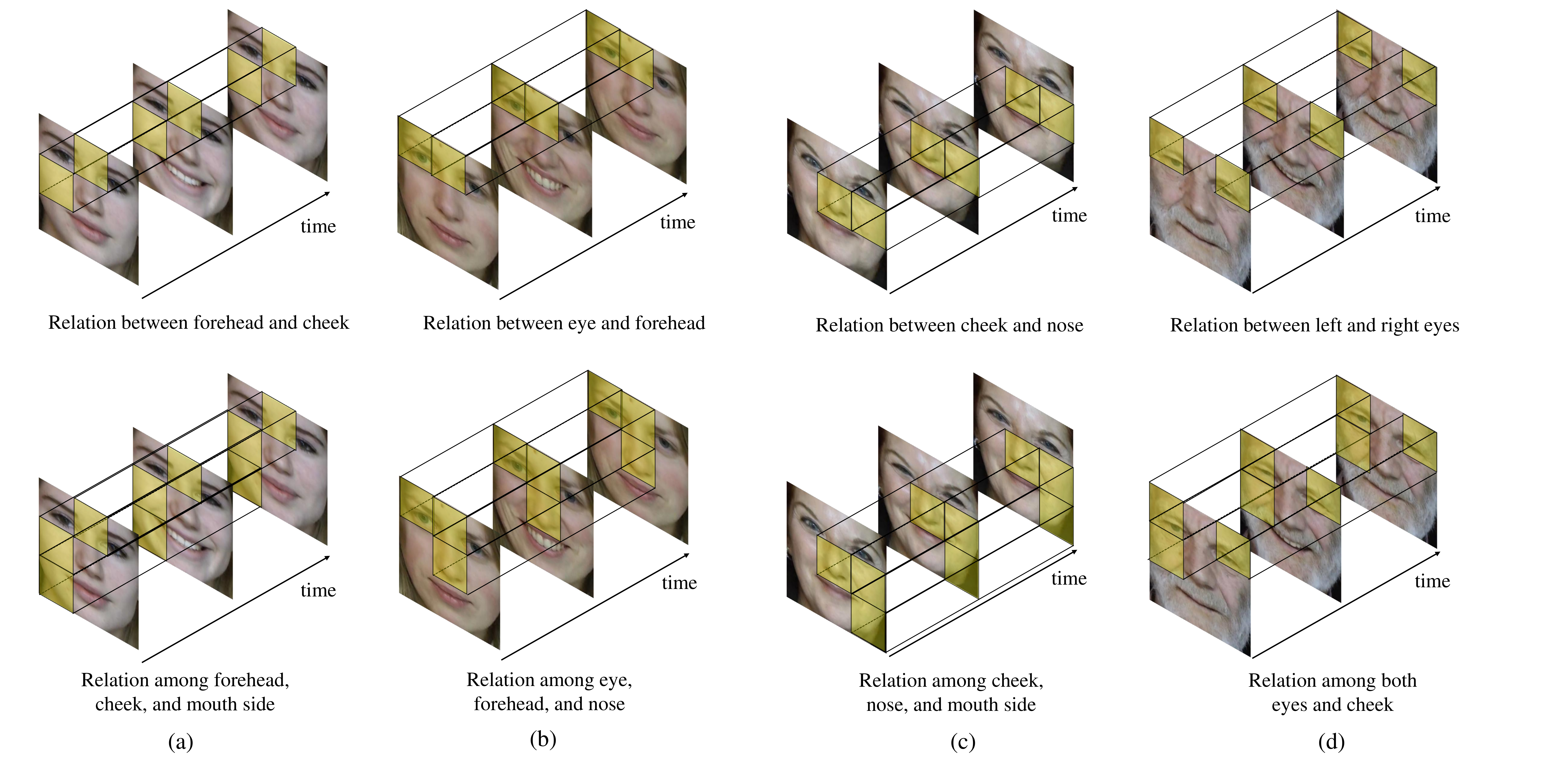}\\
	\caption{Example of facial dynamic interpretation in age estimation. Most important relations are visualized with yellow box for the relation between 2 objects in upper side and the relation among 3 objects in bottom side. (a) age group of [13-19], (b) age group of [20-36], (c) age group of [37-66], (d) age group of 66+.}\label{fig2}
\end{figure}

To understand the mechanism of the proposed facial dynamics interpreter network in age estimation, the relational importance calculated from each sequence was analyzed. Figure \ref{fig2} shows the important relations where the corresponding pair has high relational importance values. We showed the difference of important regions over different ages by presenting the important relations over age groups. Ages were divided into five age groups (8-12, 13-19, 20-36, 37-65, and 66+) according to \cite{gallagher2009understanding}. To interpret the important relations between each age group, the relational importance values encoded from test set were averaged in each age group, respectively. Four groups were visualized with example face images (there was no subject to be permitted for reporting in age group of [8-12]). As shown in the figure, when estimating age group of [66+], the relation between two eye regions was important. The relation between two eye regions could represent discriminative dynamic features according to crow’s feet and sunken eyes, which could be important factors for estimating ages of the older people. In addition, when considering three objects, the relation among left eye, right eye, and left cheek had highest relational importance in age group of [66+]. There was a tendency to symmetry about the relational importance. For example, the relation among left eye, right eye, and right cheek was included in top-5 high relational importance among 84 relations in age group of [66+]. Although the relation of action unit (AU) for determining specific facial expressions has been reported \cite{ekman2002facial}, the relation of the motions for estimating age or classifying gender was not investigated. In this study, the facial dynamics interpreter network was designed to interpret the relation of motions in facial trait estimation. It was found that the relation of dynamic features related with AU 2 and AU 6 was highly used by the deep network for estimating ages in range [66+].

\begin{figure}[!t]
	\centering
	\includegraphics[width=4.7in]{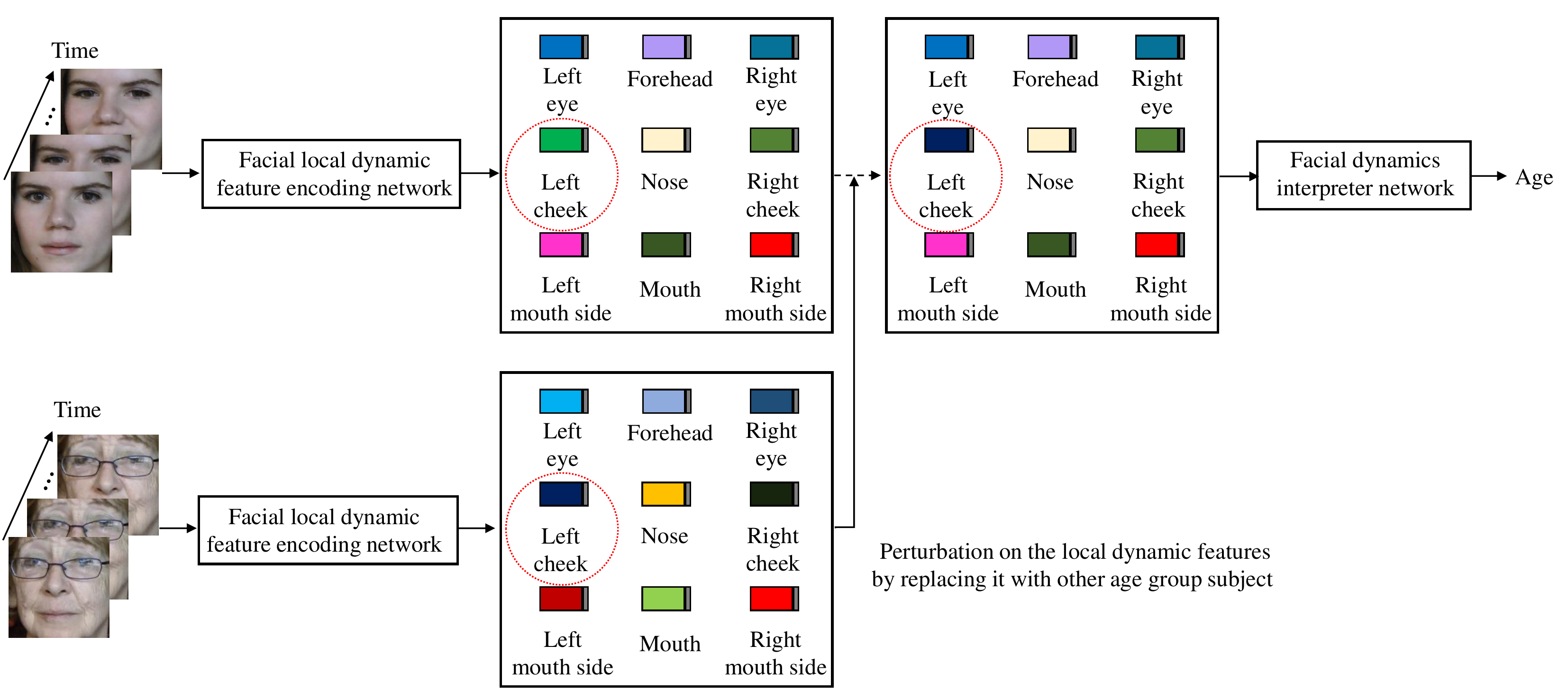}\\
	\caption{Perturbing the local dynamic features by replacing it with the local dynamic features of another age group subject (\textit{e.g.}, older subject).}\label{fig3}
	
\end{figure}

In addition, to verify the effect of important relations, we made perturbation on the dynamic features as shown in Fig.3. For the sequence of 17 years old subject, we changed the local dynamic features of left cheek region with that of 73 years old subject in the experiment. Note that the cheek constructed important pairs for estimating age group of [13-19] as shown in Fig.\ref{fig2} (a). By the perturbation, the absolute error was changed from 0.41 to 2.38. In the same way, we changed the dynamic features of other two regions (left eye and right eye) one by one. The other two regions constructed relatively less important relations and achieved the absolute error of 1.40 and 1.81 (left eye and right eye, respectively). The increase of absolute errors was less than the case which made perturbation on the left cheek. It showed that the relations with the left cheek were important for estimating age compared with the relations with eye in age group of [13-19]. 

\begin{figure}[!t]
	\centering
	\includegraphics[width=4.8in]{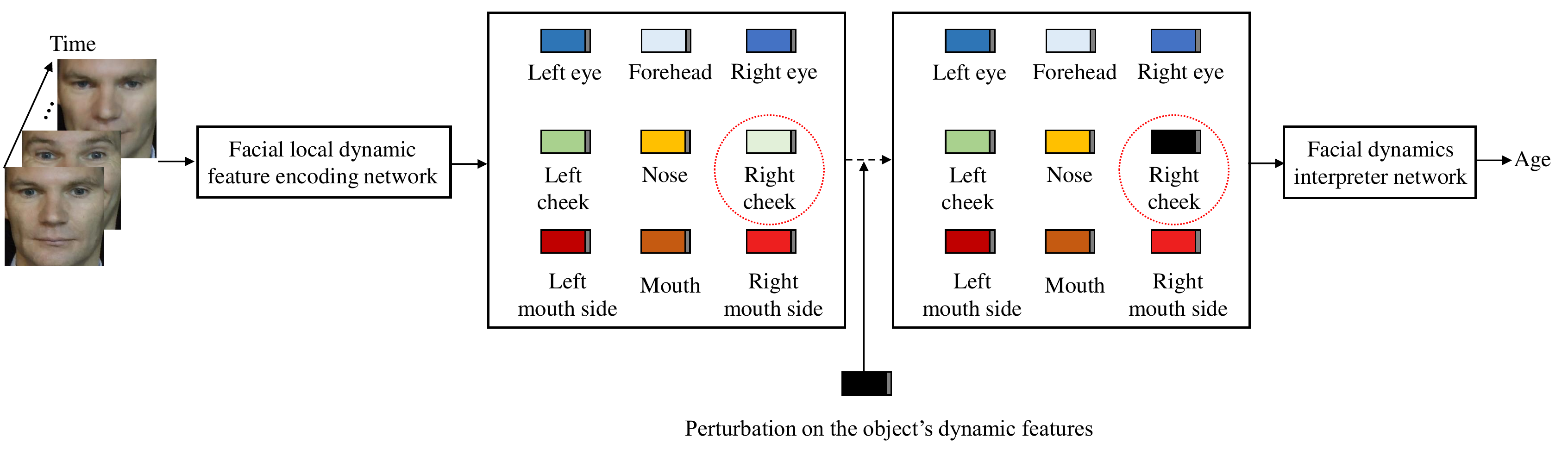}\\
	\caption{Perturbing local dynamic features by replacing it with zero vector.}\label{fig4}
\end{figure}

For the same sequence, the facial dynamics interpreter network without the use of relational importance was also analyzed. For the facial dynamics interpreter network without the use of relational importance, the absolute error of the estimated age was increased by perturbation on the local dynamic feature of the left cheek from 1.20 to 7.45. When conducting perturbation on the left eye and the right eye, the absolute errors were 1.87 and 4.21, respectively. The increase of absolute error became much larger when conducting perturbation on the left cheek. Moreover, the increase of error was larger when the facial dynamics interpreter network did not use relational importance. In other words, the facial dynamics interpreter network with the relational importance was more robust to feature contamination because it adaptively encoded the relational importance from the relational features as in Eq. (4). 

\begin{table}[!t]
	\begin{center}
		\caption{Mean absolute errors (MAE) measured after perturbing local dynamic features of different location for subjects with age groups of [37-66].}
		\label{table1}
		\begin{tabular}{l c}
			\hline
			\textbf{Perturbation location} & \textbf{MAE (standard error)}\\
			\hline	
			\hline
			Not contaminated &  5.05 ($\pm$0.28)   \\
			Less important parts (left eye, forehead, right eye) &  7.56 ($\pm$0.35)\\
			Important part (right cheek) &   9.00 ($\pm$0.41) \\
			\hline	
		\end{tabular}
	\end{center}
\end{table}

In order to statistically analyze the effect of contaminated features in the proposed facial dynamics interpreter network, we also evaluated the MAE when conducting perturbation on each dynamic features of facial local parts with zero vector as shown in Fig.~\ref{fig4}. For 402 videos which were collected from the subjects in age group of [37-66] in the UvA-NEMO database, the MAE was calculated as shown in Table 1. As shown in the table, the perturbation on most important facial region (\textit{i.e.} right cheek in age group of [37-66]) had more influenced the accuracy of age estimation compared with the case which made perturbation on less important parts (\textit{i.e.} left eye, forehead, and right eye in age group of [37-66]). The difference of MAE between the cases which made perturbation on important part and less important parts was statically significant (\textit{p}$<$0.05).

\begin{table}[t]
	\begin{center}
		\caption{Mean absolute error (MAE) of age estimation on UvA-NEMO smile database for analyzing the effectiveness of locational features and relational importance. L.F. and R.I. denote locational features and relational importance, respectively.}
		\label{table2}
		
		\begin{tabular}{l c c c c c c c}
			
			\hline
			\multicolumn{3}{l}{\textbf{Method}}& \textbf{MAE (years)}  \\
			\hline
			& & &  Posed & Spontaneous & All \\
			\hline
			\multicolumn{3}{l}{\multirow{3}{58mm}{\textbf{Aggregation of local dynamic features using regional importance}}}&& & & &\\
			& & & 4.27& 4.25 &4.26\\
			& & & & &\\
			\hline
			& \textbf{L.F.}& \textbf{R.I.} & \\
			\hhline{~--~}
			\multirow{3}{35mm}{\textbf{Facial dynamics interpreter network}}&  &  &4.05 & 4.07 & 4.06\\
			& \ding{52} &  &3.95&4.05&4.00\\
			& \ding{52} & \ding{52} &\textbf{3.83}&\textbf{3.90}&\textbf{3.87}\\
			\hline
		\end{tabular}	
	\end{center}
\end{table}

\subsubsection{Assessment of Facial Dynamics Interpreter Network for Age Estimation.}
We evaluated the effectiveness of the facial dynamics interpreter network. First, the effects of relational importance and locational features were validated for age estimation. Table 2 shows the MAE of the facial dynamics interpreter network with locational feature and relational importance. To verify the effectiveness of the relational features, the aggregation of local dynamic features using regional importance were compared. In the aggregation of local dynamic features using regional importance approach, facial local dynamic features were aggregated with regional importance in unsupervised way. As shown in the table, using the relational features improved the accuracy of age estimation. Moreover, the locational features could improve the performance of the age estimation by making the network know the location information of the object pairs. The locational features of the objects were meaningful as the objects of the face sequence were automatically aligned by the facial landmark detection. By utilizing both the relational importance and the locational features, the proposed facial dynamics interpreter network achieved the lowest MAE of 3.87 over all test set. It was mainly due to the reason that the importance of relations for age estimation was different. By considering the importance of the relational features, the accuracy of age estimation was improved. Moreover, we further analyzed the MAE of the age estimation according to the spontaneity of the smile expression. The MAE of the facial dynamic interpreter network was slightly lower in posed smile (\textit{p}$>$0.05).

\begin{table}[t]
	\begin{center}
		\caption{ Mean absolute error (MAE) on the UvA-NEMO smile database compared with other methods.}
		\label{table3}
		\begin{tabular}{l c}
			\hline
			\textbf{Method} & \textbf{\textbf{MAE (years)}} \\
			\hline
			\hline	
			VLBP \cite{hadid2011analyzing} &  15.70   \\
			Displacement \cite{dibekliouglu2012smile} & 11.54  \\
			BIF \cite{guo2009human} & 5.78 \\
			BIF + Dynamics \cite{dibekliouglu2015combining}  & 5.03  \\
			IEF \cite{alnajar2012learning} & 4.86  \\
			IEF + Dynamics \cite{dibekliouglu2015combining} & 4.33   \\	
			Holistic dynamic approach &  4.02  \\
			Proposed method &  \textbf{3.87} \\
			\hline	
		\end{tabular}
	\end{center}
\end{table}

\begin{figure}[!t]
	\centering
	\includegraphics[width=4.8in]{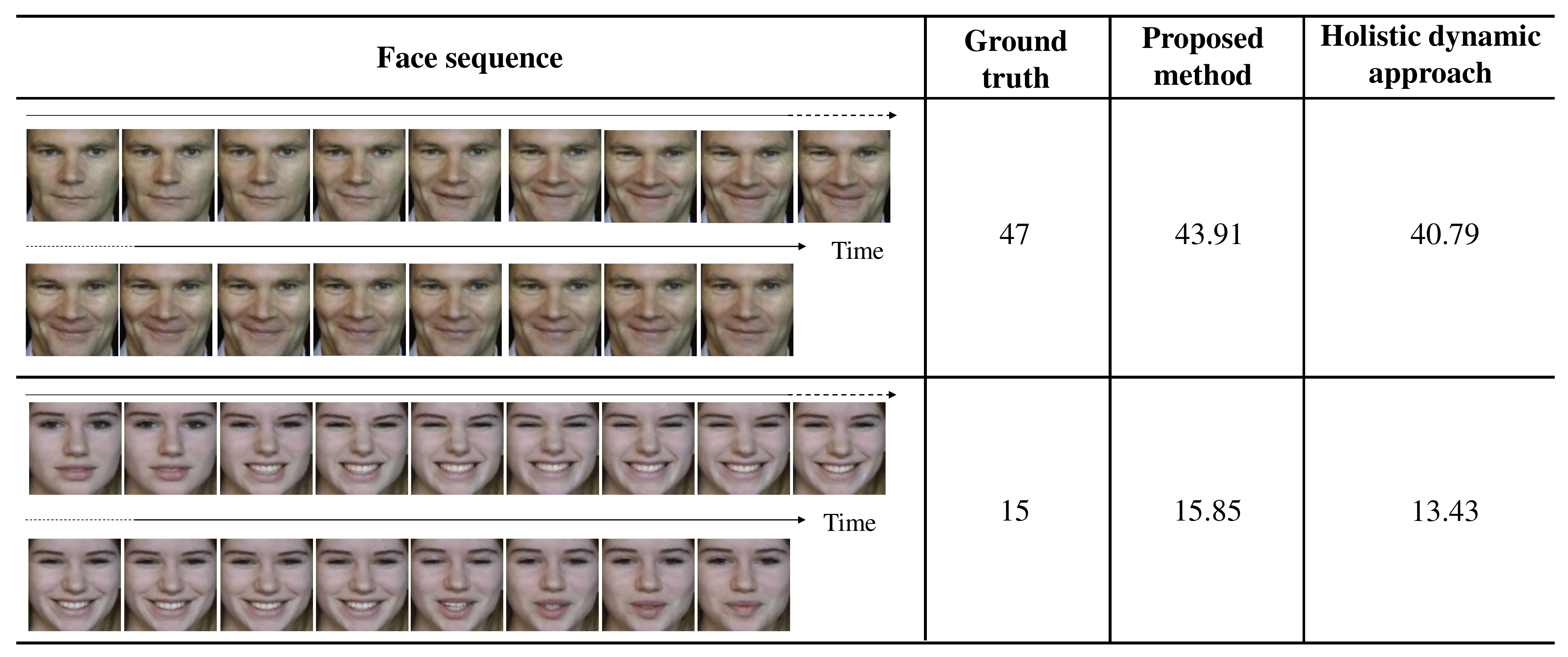}\\
	\caption{Examples of the proposed method on age estimation. For visualization purpose, face sequences are displayed in 5 frames per sec.}\label{fig5}
\end{figure}

To assess the effectiveness of the proposed dynamics interpreter network (with locational features and relational importance), the MAE of the proposed method was compared with the state-of-the-art methods (please see Table 3). The VLBP \cite{hadid2011analyzing}, displacement \cite{dibekliouglu2012smile}, BIF \cite{guo2009human}, BIF with dynamics \cite{dibekliouglu2015combining}, IEF \cite{alnajar2012learning}, IEF with dynamics \cite{dibekliouglu2015combining}, and holistic dynamic approach were compared. 
In the holistic dynamic approach, appearance features were extracted by the same VGG-face network used in the proposed method and the dynamic features were encoded with the LSTM network on the holistic appearance feature without dividing the face into local parts. It was compared because it has been widely used architecture for a spatio-temporal encoding \cite{donahue2015long,kim2017multi,kim2016facial}
As shown in the table, the proposed method achieved lowest MAE. The MAE of the proposed facial dynamics interpreter network was lower than the MAE of the IEF + Dynamics and the difference was statistically significant (\textit{p}$<$0.05). It was mainly attributed to the fact that the proposed method encoded the latent relational features from object features (facial local dynamic features and locational features) and effectively combined the relational features based on the relational importance. Examples of age estimation from the proposed method and the holistic dynamic approach are shown in Fig. 5.

\subsection{Gender Classification}

\begin{figure}[!t]
	\centering
	\includegraphics[width=3.4in]{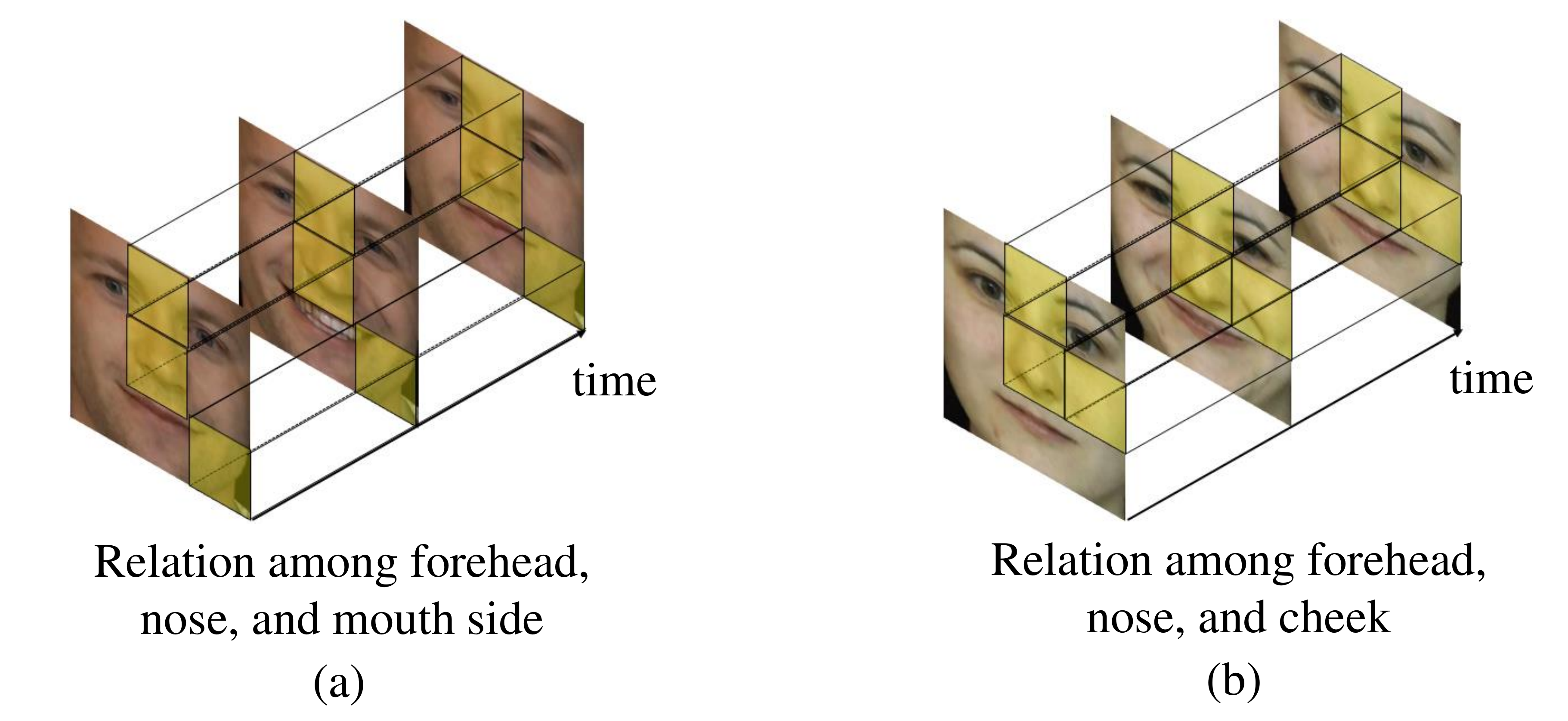}\\
	\caption{Example of interpreting important relations between facial dynamics in gender classification. Most important relations are visualized with yellow box for the relation between 3 objects for recognizing male (a) and for recognizing female (b).}\label{fig6}
\end{figure}

\subsubsection{Interpreting Relations between Facial Local Dynamics in Gender Classification.}
\label{sect:Interpreting Relations between Facial Dynamics in Gender Classification}

In order to interpret important relations in gender classification, the relational importance values encoded from each sequence were analyzed. Fig.~\ref{fig6} shows the important relations where the relational importance had high values at classifying gender from face sequence. As shown in the figure, the relation among forehead, nose, and mouth side was important in making decisions on males. Note that there was a tendency to symmetry about the relational importance. For determining male, the relation among forehead, nose, and right mouth side and the relation among forehead, nose, and left mouth side were top-2 important relations among 84 relations of three objects. For the case of female, the relation among forehead, nose, and cheek was important. It could be related to the observation that the females express emotions more frequently compared with males and the males have a tendency to show restricted emotions compared with the females. In other words, the females have a tendency to make smiles bigger than males by using muscles of cheek regions. Therefore, the relations among cheek and other face parts were important for recognizing females.

\begin{table}[t]
	\begin{center}
		\caption{Accuracy of gender classification on the UvA-NEMO smile database to analyze the effectiveness of the locational feature and relational importance. L.F. and R.I. denote locational features and relational importance, respectively.}
		\label{table4}
		\begin{tabular}{l c c c}
			\hline
			\multicolumn{3}{l}{\textbf{Method}}& \multirow{2}{25mm}{\textbf{Classification Accuracy ($\%$)}}  \\
			&\\
			\hline
			\hline
			\multicolumn{3}{l}{Aggregation of local dynamic features using regional importance} &88.87\\
			\hline
			& \textbf{L.F.}& \textbf{R.I.} & \\
			\hhline{~--~}
			&  &  &88.79\\
			{Facial dynamics interpreter network} & \ding{52} &  &89.35\\
			& \ding{52} & \ding{52} &\textbf{90.08}\\
			\hline
		\end{tabular}
	\end{center}
\end{table}

\subsubsection{Assessment of Facial Dynamics Interpreter Network for Gender Classification.}

We also evaluated the effectiveness of the proposed facial dynamics interpreter network for gender classification. First, the classification accuracy of the facial dynamics interpreter network with relational importance and locational features are summarized in Table 4. For comparison, the aggregation of local dynamic features using regional importance was compared. The proposed facial dynamics interpreter network achieved the highest accuracy by using both locational features and relational importance. The locational features and the relational importance in the facial dynamics interpreter network were also important for gender classification. 

\begin{table}[t]
	\begin{center}
		\caption{Accuracy of gender classification on the UvA-NEMO smile database compared with other methods.}
		\label{table5}
		\begin{tabular}{l c c c c c c}
			\hline
			\multirow{3}{*}{\textbf{Method}} & \multicolumn{6}{c}{\textbf{Classification Accuracy ($\%$)}} \\
			\hhline{~------}
			&\multicolumn{2}{l}{Spontaneous} & &\multicolumn{2}{l}{Posed} & \multirow{2}{*}{All}\\
			\hhline{~-----}
			& Age$<=$19 & Age$>$19 & &Age$<=$19 & Age$>$19 &  \\
			\hline
			\hline
			\textit{how-old.net} + dynamics(Tree) \cite{dantcheva2017gender} & 60.80 & 93.46 && N/A & N/A & N/A\\
			\textit{how-old.net} + dynamics(SVM) \cite{dantcheva2017gender} & N/A & N/A && 60.80 & 92.89 & N/A\\
			COTS + dynamics (Tree) \cite{dantcheva2017gender} & 76.92 & 93.00 && N/A & N/A & N/A \\
			COTS + dynamics (Bagged Trees) \cite{dantcheva2017gender} & N/A & N/A && 76.92 & 92.89 & N/A \\
			Image-based CNN \cite{levi2015age} & 80.72 & 89.85 && 80.58 & 92.36 & 86.94 \\
			Holistic dynamic approach & 74.38 & 93.52 && 77.51 & 93.91 & 87.10 \\
			Proposed method & 80.17& 94.65&& 85.14 & 95.18& \textbf{90.08}\\
			\hline
		\end{tabular}
	\end{center}
\end{table}

Table~\ref{table5} shows the classification accuracy of the proposed facial dynamics interpreter network compared with other methods on UvA-NEMO database. Two types of appearance based approach named “how-old.net” and “commercial off-the-shelf (COTS)” were combined with a hand-crafted dynamic approach for gender classification \cite{dantcheva2017gender}. How-old.net was a website (http://how-old.net/) launched by Microsoft for online age and gender recognition. COTS was a commercial face detection and recognition software, which included a gender classification. The dynamic approach calculated the facial local region’s dynamic descriptors such as amplitude, speed, and acceleration  as described in \cite{dantcheva2017gender}. In holistic dynamic approach, appearance features were extracted by the same VGG-face network used in the proposed method and the dynamic features were encoded on the holistic appearance features. An image based method \cite{levi2015age} was also compared to validate the effectiveness of utilizing facial dynamics in gender classification. The accuracy of how-old.net+dynamics and COTS+dynamics were directly from \cite{dantcheva2017gender} and the accuracy of the image-based CNN and the holistic dynamic approach were calculated in this study. By exploiting the relations between local dynamic features, the proposed method achieved the highest accuracy compared with other methods. The performance difference between the holistic approach and the proposed method was statistically significant (\textit{p}$<$0.05).

\section{Conclusions}

According to cognitive-psychological studies, facial dynamics could provide crucial cues for face analysis. The motion of facial local regions from facial expression is known that it is related to the motion of other facial regions. In this paper, the novel deep learning approach which could interpret the relations between facial local dynamics was proposed to interpret relations of local dynamics in facial trait estimation from the smile expression. Facial traits were estimated by combining relational features of facial local dynamics based on the relational importance. By comparative experiments, the effectiveness of the proposed method was verified for facial trait estimation. The important relations between facial dynamics were interpreted by the proposed method in gender classification and age estimation. The proposed method could accurately estimate facial traits (age and gender) compared with the state-of-the-art methods. We will attempt to extend the proposed method to other facial dynamic analysis such as spontaneity analysis \cite{dibekliouglu2012you} and video facial expression recognition \cite{kim2017multi}.

\subsubsection{Acknowledgement.} This work was partly supported by Institute for Information \& Communications Technology Promotion (IITP) grant funded by the Korea government (MSIT) (No. 2017-0-01778, Development of Explainable Human-level Deep Machine Learning Inference Framework) and (No. 2017-0-00111, Practical Technology Development of High Performing Emotion Recognition and Facial Expression based Authentication using Deep Network). {*}Y.M. Ro is a corresponding author.

\clearpage
\bibliographystyle{splncs04}
\bibliography{egbib}

\begin{thebibliography}{10}
\providecommand{\url}[1]{\texttt{#1}}
\providecommand{\urlprefix}{URL }
\providecommand{\doi}[1]{https://doi.org/#1}

\bibitem{adams2015intersection}
Adams~Jr, R.B., Hess, U., Kleck, R.E.: The intersection of gender-related
  facial appearance and facial displays of emotion. Emotion Review
  \textbf{7}(1),  5--13 (2015)

\bibitem{alnajar2012learning}
Alnajar, F., Shan, C., Gevers, T., Geusebroek, J.M.: Learning-based encoding
  with soft assignment for age estimation under unconstrained imaging
  conditions. Image and Vision Computing  \textbf{30}(12),  946--953 (2012)

\bibitem{asthana2014incremental}
Asthana, A., Zafeiriou, S., Cheng, S., Pantic, M.: Incremental face alignment
  in the wild. In: Proceedings of the IEEE Conference on Computer Vision and
  Pattern Recognition. pp. 1859--1866 (2014)

\bibitem{bekios2014robust}
Bekios-Calfa, J., Buenaposada, J.M., Baumela, L.: Robust gender recognition by
  exploiting facial attributes dependencies. Pattern Recognition Letters
  \textbf{36},  228--234 (2014)

\bibitem{cashdan1998smiles}
Cashdan, E.: Smiles, speech, and body posture: How women and men display
  sociometric status and power. Journal of Nonverbal Behavior  \textbf{22}(4),
  209--228 (1998)

\bibitem{dantcheva2017gender}
Dantcheva, A., Br{\'e}mond, F.: Gender estimation based on smile-dynamics. IEEE
  Transactions on Information Forensics and Security  \textbf{12}(3),  719--729
  (2017)

\bibitem{dantcheva2016else}
Dantcheva, A., Elia, P., Ross, A.: What else does your biometric data reveal? a
  survey on soft biometrics. IEEE Transactions on Information Forensics and
  Security  \textbf{11}(3),  441--467 (2016)

\bibitem{demirkus2010gender}
Demirkus, M., Toews, M., Clark, J.J., Arbel, T.: Gender classification from
  unconstrained video sequences. In: Computer Vision and Pattern Recognition
  Workshops (CVPRW), 2010 IEEE Computer Society Conference on. pp. 55--62. IEEE
  (2010)

\bibitem{dibekliouglu2015combining}
Dibeklio{\u{g}}lu, H., Alnajar, F., Salah, A.A., Gevers, T.: Combining facial
  dynamics with appearance for age estimation. IEEE Transactions on Image
  Processing  \textbf{24}(6),  1928--1943 (2015)

\bibitem{dibekliouglu2012smile}
Dibeklio{\u{g}}lu, H., Gevers, T., Salah, A.A., Valenti, R.: A smile can reveal
  your age: Enabling facial dynamics in age estimation. In: Proceedings of the
  20th ACM international conference on Multimedia. pp. 209--218. ACM (2012)

\bibitem{dibekliouglu2012you}
Dibeklio{\u{g}}lu, H., Salah, A.A., Gevers, T.: Are you really smiling at me?
  spontaneous versus posed enjoyment smiles. In: European Conference on
  Computer Vision. pp. 525--538. Springer (2012)

\bibitem{dibekliouglu2015recognition}
Dibeklio{\u{g}}lu, H., Salah, A.A., Gevers, T.: Recognition of genuine smiles.
  IEEE Transactions on Multimedia  \textbf{17}(3),  279--294 (2015)

\bibitem{donahue2015long}
Donahue, J., Anne~Hendricks, L., Guadarrama, S., Rohrbach, M., Venugopalan, S.,
  Saenko, K., Darrell, T.: Long-term recurrent convolutional networks for
  visual recognition and description. In: Proceedings of the IEEE conference on
  computer vision and pattern recognition. pp. 2625--2634 (2015)

\bibitem{ekman2002facial}
Ekman, P.: Facial action coding system (facs). A human face  (2002)

\bibitem{gallagher2009understanding}
Gallagher, A.C., Chen, T.: Understanding images of groups of people. In:
  Computer Vision and Pattern Recognition, 2009. CVPR 2009. IEEE Conference on.
  pp. 256--263. IEEE (2009)

\bibitem{guo2009human}
Guo, G., Mu, G., Fu, Y., Huang, T.S.: Human age estimation using bio-inspired
  features. In: Computer Vision and Pattern Recognition, 2009. CVPR 2009. IEEE
  Conference on. pp. 112--119. IEEE (2009)

\bibitem{hadid2011analyzing}
Hadid, A.: Analyzing facial behavioral features from videos. Human Behavior
  Understanding pp. 52--61 (2011)

\bibitem{hess2004facial}
Hess, U., Adams~Jr, R.B., Kleck, R.E.: Facial appearance, gender, and emotion
  expression. Emotion  \textbf{4}(4), ~378 (2004)

\bibitem{hochreiter1997long}
Hochreiter, S., Schmidhuber, J.: Long short-term memory. Neural computation
  \textbf{9}(8),  1735--1780 (1997)

\bibitem{ioffe2015batch}
Ioffe, S., Szegedy, C.: Batch normalization: Accelerating deep network training
  by reducing internal covariate shift. In: International Conference on Machine
  Learning. pp. 448--456 (2015)

\bibitem{ji20133d}
Ji, S., Xu, W., Yang, M., Yu, K.: 3d convolutional neural networks for human
  action recognition. IEEE transactions on pattern analysis and machine
  intelligence  \textbf{35}(1),  221--231 (2013)

\bibitem{juefei2016deepgender}
Juefei-Xu, F., Verma, E., Goel, P., Cherodian, A., Savvides, M.: Deepgender:
  Occlusion and low resolution robust facial gender classification via
  progressively trained convolutional neural networks with attention. In:
  Proceedings of the IEEE Conference on Computer Vision and Pattern Recognition
  Workshops. pp. 68--77 (2016)

\bibitem{karpathy2014large}
Karpathy, A., Toderici, G., Shetty, S., Leung, T., Sukthankar, R., Fei-Fei, L.:
  Large-scale video classification with convolutional neural networks. In:
  Proceedings of the IEEE conference on Computer Vision and Pattern
  Recognition. pp. 1725--1732 (2014)

\bibitem{kim2017multi}
Kim, D.H., Baddar, W., Jang, J., Ro, Y.M.: Multi-objective based
  spatio-temporal feature representation learning robust to expression
  intensity variations for facial expression recognition. IEEE Transactions on
  Affective Computing  (2017)

\bibitem{kim2016facial}
Kim, S.T., Kim, D.H., Ro, Y.M.: Facial dynamic modelling using long short-term
  memory network: Analysis and application to face authentication. In:
  Biometrics Theory, Applications and Systems (BTAS), 2016 IEEE 8th
  International Conference on. pp.~1--6. IEEE (2016)

\bibitem{lea2016segmental}
Lea, C., Reiter, A., Vidal, R., Hager, G.D.: Segmental spatiotemporal cnns for
  fine-grained action segmentation. In: European Conference on Computer Vision.
  pp. 36--52. Springer (2016)

\bibitem{levi2015age}
Levi, G., Hassner, T.: Age and gender classification using convolutional neural
  networks. In: Proceedings of the IEEE Conference on Computer Vision and
  Pattern Recognition Workshops. pp. 34--42 (2015)

\bibitem{li2015shape}
Li, S., Xing, J., Niu, Z., Shan, S., Yan, S.: Shape driven kernel adaptation in
  convolutional neural network for robust facial traits recognition. In:
  Proceedings of the IEEE Conference on Computer Vision and Pattern
  Recognition. pp. 222--230 (2015)

\bibitem{makinen2008evaluation}
Makinen, E., Raisamo, R.: Evaluation of gender classification methods with
  automatically detected and aligned faces. IEEE Transactions on Pattern
  Analysis and Machine Intelligence  \textbf{30}(3),  541--547 (2008)

\bibitem{nair2010rectified}
Nair, V., Hinton, G.E.: Rectified linear units improve restricted boltzmann
  machines. In: Proceedings of the 27th international conference on machine
  learning (ICML-10). pp. 807--814 (2010)

\bibitem{parkhi2015deep}
Parkhi, O.M., Vedaldi, A., Zisserman, A., et~al.: Deep face recognition. In:
  BMVC. vol.~1, p.~6 (2015)

\bibitem{pilz2006search}
Pilz, K.S., Thornton, I.M., B{\"u}lthoff, H.H.: A search advantage for faces
  learned in motion. Experimental Brain Research  \textbf{171}(4),  436--447
  (2006)

\bibitem{reid2013soft}
Reid, D., Samangooei, S., Chen, C., Nixon, M., Ross, A.: Soft biometrics for
  surveillance: an overview. Machine learning: theory and applications.
  Elsevier pp. 327--352 (2013)

\bibitem{roark2003psychological}
Roark, D.A., Barrett, S.E., Spence, M.J., Abdi, H., O'Toole, A.J.:
  Psychological and neural perspectives on the role of motion in face
  recognition. Behavioral and cognitive neuroscience reviews  \textbf{2}(1),
  15--46 (2003)

\bibitem{santoro2017simple}
Santoro, A., Raposo, D., Barrett, D.G., Malinowski, M., Pascanu, R., Battaglia,
  P., Lillicrap, T.: A simple neural network module for relational reasoning.
  In: Advances in neural information processing systems. pp. 4974--4983 (2017)

\bibitem{simon2004gender}
Simon, R.W., Nath, L.E.: Gender and emotion in the united states: Do men and
  women differ in self-reports of feelings and expressive behavior? American
  journal of sociology  \textbf{109}(5),  1137--1176 (2004)

\bibitem{srivastava2014dropout}
Srivastava, N., Hinton, G.E., Krizhevsky, A., Sutskever, I., Salakhutdinov, R.:
  Dropout: a simple way to prevent neural networks from overfitting. Journal of
  machine learning research  \textbf{15}(1),  1929--1958 (2014)

\bibitem{toews2009detection}
Toews, M., Arbel, T.: Detection, localization, and sex classification of faces
  from arbitrary viewpoints and under occlusion. IEEE Transactions on Pattern
  Analysis and Machine Intelligence  \textbf{31}(9),  1567--1581 (2009)

\bibitem{tong2007facial}
Tong, Y., Liao, W., Ji, Q.: Facial action unit recognition by exploiting their
  dynamic and semantic relationships. IEEE transactions on pattern analysis and
  machine intelligence  \textbf{29}(10) (2007)

\bibitem{tran2015learning}
Tran, D., Bourdev, L., Fergus, R., Torresani, L., Paluri, M.: Learning
  spatiotemporal features with 3d convolutional networks. In: Proceedings of
  the IEEE international conference on computer vision. pp. 4489--4497 (2015)

\bibitem{uvrivcavr2016structured}
U{\v{r}}i{\v{c}}a{\v{r}}, M., Timofte, R., Rothe, R., Matas, J., et~al.:
  Structured output svm prediction of apparent age, gender and smile from deep
  features. In: Proceedings of the 29th IEEE Conference on Computer Vision and
  Pattern Recognision Workshop (CVPRW 2016). pp. 730--738. IEEE (2016)

\bibitem{yue2015beyond}
Yue-Hei~Ng, J., Hausknecht, M., Vijayanarasimhan, S., Vinyals, O., Monga, R.,
  Toderici, G.: Beyond short snippets: Deep networks for video classification.
  In: Proceedings of the IEEE conference on computer vision and pattern
  recognition. pp. 4694--4702 (2015)

\bibitem{zhao2016joint}
Zhao, K., Chu, W.S., De~la Torre, F., Cohn, J.F., Zhang, H.: Joint patch and
  multi-label learning for facial action unit and holistic expression
  recognition. IEEE Transactions on Image Processing  \textbf{25}(8),
  3931--3946 (2016)

\end{thebibliography}

\end{document}